\documentclass[11pt]{article}

\usepackage[final]{acl}

 \usepackage{microtype}
\usepackage{enumitem}
\usepackage[english,bidi=default]{babel} 
\babelfont{rm}{TeXGyreTermesX} 
\babelprovide[import]{hindi}
\babelfont[*devanagari]{rm}{Lohit Devanagari}
\babelprovide[import]{arabic}
\babelfont[*arabic]{rm}{Noto Sans Arabic}

\usepackage{graphicx}
\usepackage{caption}
\usepackage{comment}
\usepackage{amsmath}
\usepackage{algorithm}
\usepackage{algorithmic}
\newcommand{\concat}{\mathbin{\|}}
\usepackage{multirow}
\usepackage{booktabs}
\DeclareMathOperator*{\argmax}{arg\,max}

\usepackage{subcaption}
\usepackage{xspace}
\newcommand{\name}{BlendIn\xspace}

\ifodd 0

\newcommand{\revjg}[1]{{\color{red}#1}} 
   
\newcommand{\needrev}[1]{{\color{green}#1}}

\else

\newcommand{\revjg}[1]{#1}   
 
\newcommand{\needrev}[1]{#1}

\fi 

\title{To Intervene or Not: Guiding Inference-time Alignment with Probabilistic Model Blending}

\author{
    \textbf{Jin Gan},
    \textbf{Xin Li\thanks{Corresponding author.}},
    \textbf{Jun Luo}
    \\
    \\
    College of Computing and Data Science, Nanyang Technological University, Singapore
    \\
   \href{jin010@ntu.edu.sg}{\{jin010,~l.xin,~junluo\}@ntu.edu.sg}
}

\begin{document}

\maketitle

\begin{abstract}
The wide deployment of LLMs has made model alignment necessary to make newly trained models safely and effectively respond to user instructions. Among different methods, inference-time alignment is often cheaper as it intervenes (i.e., offers guidances) only during output generation. Existing proposals apply guidances extracted from certain aligned models without properly assessing their reliability. Nonetheless, our systematic evaluation reveals that guidance effectiveness varies drastically across models; since ineffective guidances lead to further confusion and thus further interventions, the resulting excessive interventions typically indicate poor performance. To make interventions more effective and thus more efficient, we introduce \name, 
an 
inference-time alignment framework that shifts from 
binary decisions 
to creating hybrid distributions integrating both models' 
knowledge. 
\name 
stabilizes inference-time alignment by performing 
quality-aware alignment and proportionally weighting 
each model's contribution based on reliability. 
Compared with existing works, it preserves beneficial guidance while downweighting 
unreliable suggestions. 
\name provides both diagnostic signals and mitigation strategies for 
misaligned guidance, achieving consistent and up to \revjg{50\%} performance improvement on challenging model pairs.
Our code is available at: \url{https://github.com/DecayingSeart/BlendIn}.
\end{abstract}

\section{Introduction}

\begin{figure*}[h]
  \setlength\abovecaptionskip{8pt}
  \centering 
  \begin{minipage}[b]{0.485\linewidth}
  \centering 
    \includegraphics[height=.5\linewidth]{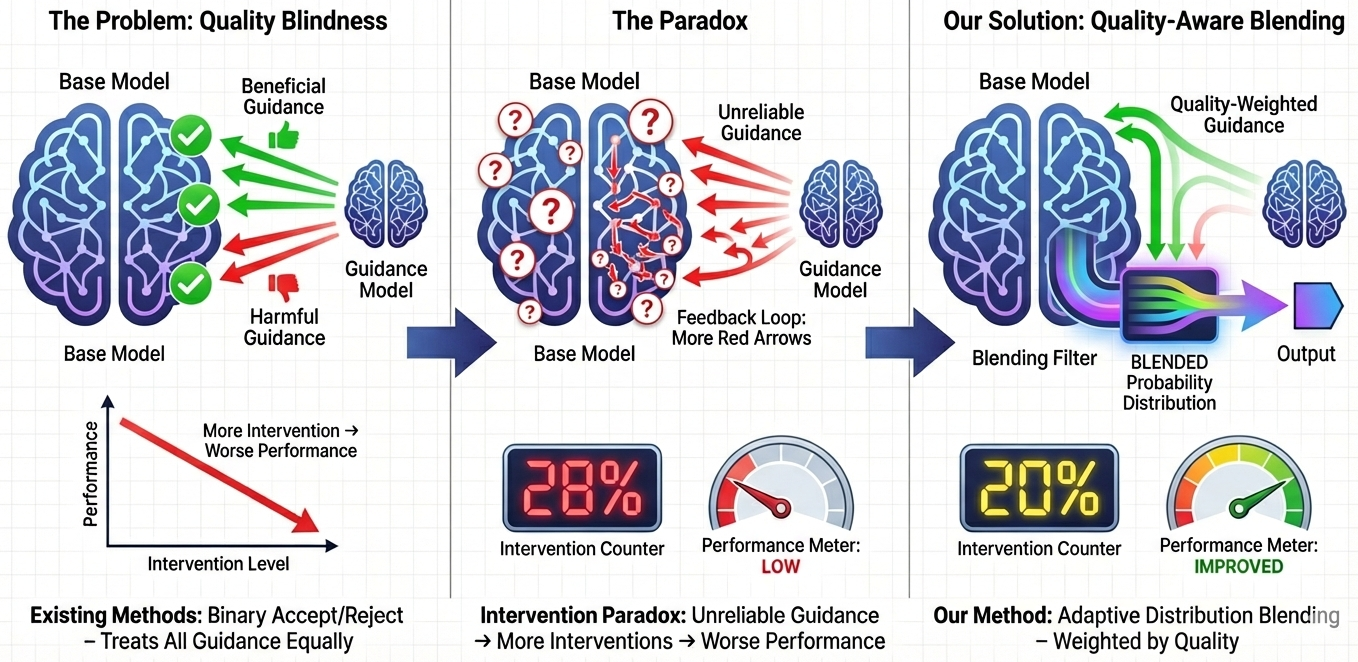}
    \subcaption{}
    \label{fig:teaser}    
  \end{minipage}
  \hfill
  \begin{minipage}[b]{0.485\linewidth}
  \centering 
    \includegraphics[height=.5\linewidth]{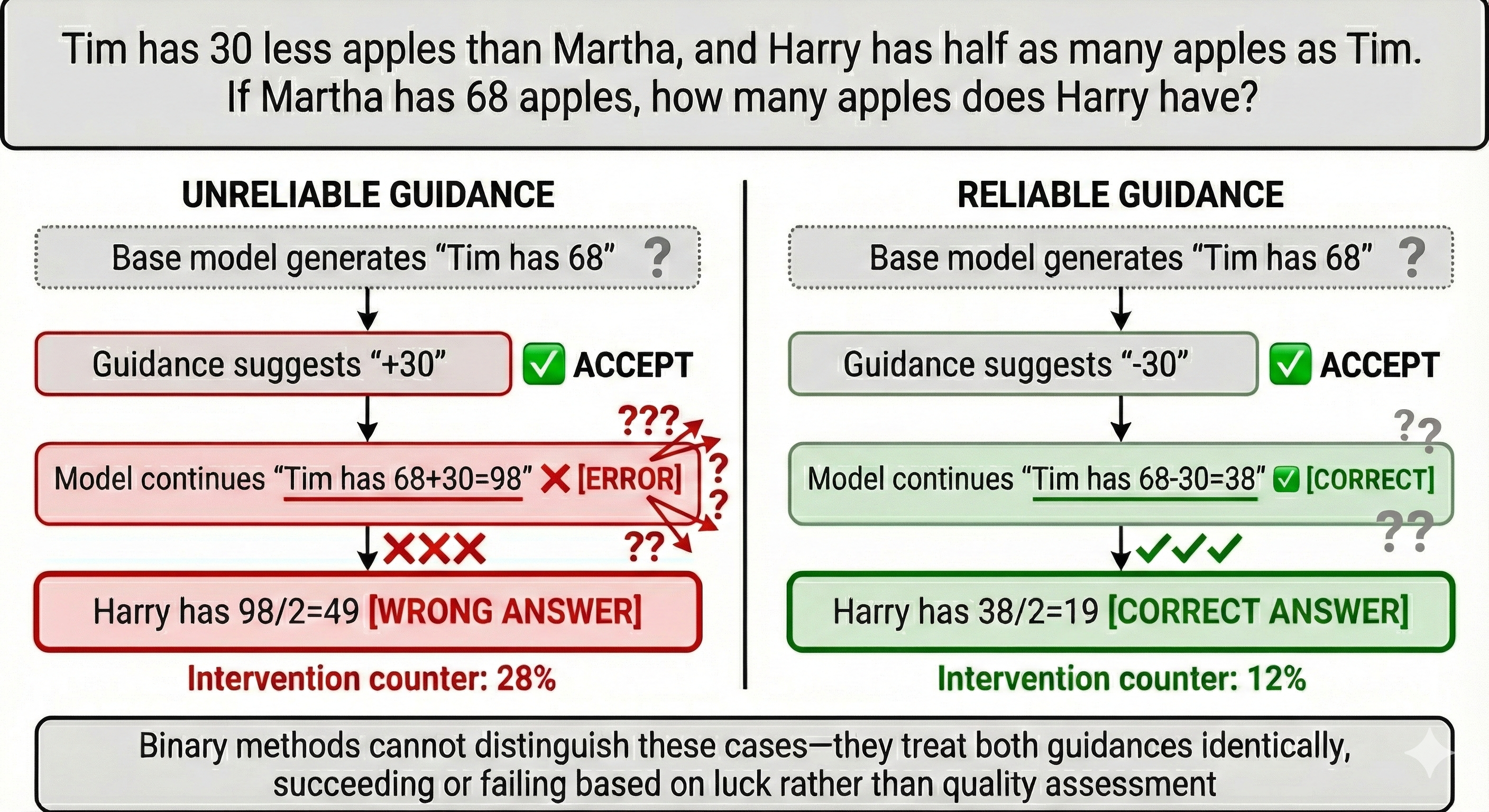}
    \subcaption{}
    \label{fig:example}   
  \end{minipage}
  \vspace{-0.5em}
  \caption{(a) Overview of quality blindness in inference-time alignment and our solution.
    \textbf{Left:} Existing methods make binary accept/reject decisions, treating all guidance 
    equally without assessing quality--- accepting both beneficial (green) and harmful (red) 
    suggestions identically. \textbf{Center:} This leads to an intervention paradox: unreliable 
    guidance triggers cascading failures that require more interventions, creating a negative 
    correlation between intervention rate and performance. \textbf{Right:} \name addresses this 
    through quality-aware distribution blending, weighting guidance proportionally to 
    reliability rather than making binary decisions. (b) Cascading failure from unreliable guidance. \textbf{Left:} When unreliable 
    guidance suggests an incorrect token ('+' instead of '-'), binary acceptance 
    propagates this error through subsequent steps, creating uncertainty that triggers 
    additional interventions (28\% intervention rate) and produces wrong answers. 
    \textbf{Right:} Reliable guidance providing correct suggestions enables accurate 
    generation with minimal intervention (12\%). Existing binary methods cannot distinguish 
    these cases, treating both guidances identically despite opposite outcomes.}
  \label{fig:final_teaser}
  \vspace{-1.7em}
\end{figure*}

\revjg{The helpfulness and safety 
of large language models (LLMs) largely depends on their alignment to follow user instructions}~\cite{nudging}. 
This is traditionally achieved through 
fine-tuning 
methods where 
alignment must be performed separately for each model newly trained
~\cite{RLHF, DPO}, incurring substantial computational costs. Such inefficiency has motivated 
inference-time alignment, 
which 
leverages aligned models or their extracted signals~\cite{nudging, IVG, inferaligner} as \textit{guidance models} to align unaligned models as \textit{base models}~\cite{inferaligner, nudging, IVG} during output generation, thereby avoiding expensive retraining. 

Existing works on inference-time alignment provides guidance in different forms of signals like token 
suggestions, value scores, or activation steering~\cite{nudging, inferaligner, IVG}. 
However, these methods lack mechanisms to assess 
if the guidance itself is reliable. Their designs implicitly assume that all guidance is beneficial--- 
an assumption that our systematic analysis refutes. Across nine models~\cite{gemma3, gemma2, llama3, qwen3} and six datasets~\cite{gsm8k, mmlu, mmlu2, lin-etal-2022-truthfulqa, arcchallenge, justeval}, we observe a dramatic variance on guidance effectiveness: 
some 
model combinations succeed while others fail catastrophically. Critically, 
model pairs 
with 
excessive intervention rates 
systematically 
perform worse, 
not better. This counterintuitive pattern reveals that guidance effectiveness 
varies fundamentally across different model combinations. 

Why does this happen? When base models encounter difficult positions in generation where 
they have misaligned or harmful prediction, guidance models may also 
struggle at these same positions--- 
they make 
incorrect predictions about what tokens should come next. 
As a result, these models provide 
misaligned or harmful suggestions, 
leading to 
misaligned 
outputs that trigger even more interventions. 
Because existing methods 
cannot distinguish 
beneficial suggestions from harmful ones, 
they exhibit a fundamental problem of quality blindness. 
Without mechanisms to 
predict or mitigate this issue, 
alignment success must rely on expensive trial-and-error 
testing, limiting practical deployment of inference-time alignment. See Figure~\ref{fig:teaser} for overview and Figure~\ref{fig:example} for example.
These observations suggest that effective inference-time alignments require 
a shift in approach. 
Existing methods' quality blindness leads to binary accept/reject 
decisions that cannot express partial trust or selective integration. 
When 
guidance is unreliable, binary methods must choose between accepting harmful 
suggestions (degrading performance) or rejecting all guidance (losing 
potential benefits). Neither 
is satisfactory.

We argue that quality-aware integration is essential: methods must assess 
guidance reliability at each intervention position and integrate both models' knowledge 
proportionally rather than making all-or-nothing decisions. This enables critical capabilities that binary methods lack. First, it preserves 
beneficial portions of guidance while downweighting unreliable portions. Second, 
it leverages partial knowledge from both models simultaneously. 
Third, it gracefully handles varying degrees of guidance reliability.

Building on these insights, we propose \name, 
a quality-aware 
inference time alignment method based on blending distributions.
At each position where base model demonstrates low confidence in prediction, \name evaluates guidance quality by blending the full probability distributions from both guidance and base models using adaptive weights based on their respective certainties. 
Rather than making binary decisions, 
we sample from the hybrid distribution using greedy selection, allowing both models 
to contribute proportionally. 
This 
soft integration effectively filters 
misaligned or harmful 
interventions while 
preserving beneficial guidance, addressing the quality blindness in 
existing approaches. 
Our approach achieves consistent and maximum 50\% improvements 
on challenging high-intervention pairs, demonstrating its effectiveness and robustness.

Our contributions include: 

\begin{itemize}[itemsep=2pt,topsep=2pt,parsep=2pt] 
\item 
Making novel contributions toward systematic characterization of quality failures in inference-time alignment, identifying the negative correlation between excessive intervention rate and overall performance. 
\item Proposing \name, 
an 
inference-time alignment stabilization method 
that diagnoses and mitigates 
unreliable guidance through soft distribution blending. 
\item Improving performance on challenging over-intervened model pairs consistently and up to 50\%.
\end{itemize}

Our work transforms inference-time alignment from an empirically promising but insufficiently predictable technique into one with principled diagnostic signals and effective 
mitigation to misaligned guidance. 
The rest of the paper is organized as follows. Section 2 briefly captures related works. Section 3 presents \revjg{problem formulation.}
Section 4 discusses details of our method. Section 5 reports the experimental results. Finally, section 6 concludes the paper. 

\section{Related Works}

Inference-time alignment methods guide unaligned models using signals from 
aligned models without parameter updates. NUDGING~\cite{nudging} uses 
speculative decoding~\cite{speculativedecoding, speculativedecoding2} where guidance models propose tokens whenever the top 1 probability of base model falls under a threshold. 
IVG~\cite{IVG} applies a value function trained on outputs of an aligned model to pick the highest-scored token candidate in base model. 
InferAligner~\cite{inferaligner} selectively shifts base model's activations using vectors 
extracted from 
aligned models whenever harmful query is detected. While these recent methods differ in mechanism (token proposals, 
value scoring, and activation modification), they share a common design principle 
of treating guidance as uniformly beneficial. However, they lack mechanisms to 
assess whether specific guidance models actually provide aligned 
suggestions for base models. 
This 
problem 
remains uncharacterized: 
can alignment success be predicted by any property that is easy to obtain, or must every model 
combination undergo expensive empirical testing on full benchmarks? Can we neutralize such guidance quality failure in advance? Without systematic 
analysis of these questions, practitioners lack diagnostic signals 
for rapid failure detection and principled strategies for failure mitigation.

\section{\revjg{Problem Formulation}} 

\subsection{Preliminary} 

Inference-time alignment could align base models using guidance models during output generation. This could be implemented through speculative decoding~\cite{speculativedecoding, speculativedecoding2}, where an aligned model proposes tokens and the unaligned model validates them based on an uncertainty threshold~\cite{nudging}. 

Formally, at each generation step $t$, the base model first checks its own confidence. Guidance is triggered only at uncertain positions:
\begin{enumerate}[itemsep=2pt,topsep=2pt,parsep=2pt]
    \item \textbf{Uncertainty check}: Compute $u = \max_w P_{M_b}(w|x_{<t})$; if $u \geq \tau$ (uncertainty threshold), select $\arg\max_w P_{M_b}(w|x_{<t})$ and continue to the next position
    \item \textbf{Proposal phase} (only when $u < \tau$): Aligned model $M_s$ generates candidate tokens $\{w_1, \ldots, w_k\}$
    \item \textbf{Acceptance} (only when $u < \tau$): Accept the guidance model's top token $\arg\max_w P_{M_s}(w|x_{<t})$ and append to sequence
    \item \textbf{Continuation}: $M_b$ continues generation until the next uncertain position
\end{enumerate}


This design is simple as it doesn't require training extra models or functions. 

\subsection{Quality Blindness} 

Existing methods apply guidance through different mechanisms, 
but they lack mechanisms to evaluate guidance quality as they treat all guidance as uniformly beneficial. 
They cannot distinguish between aligned, helpful suggestions from misaligned, harmful suggestions. 
When the base model makes misaligned or harmful prediction at a difficult position, guidance models may share the same situation. 
Both models are effectively unreliable in this case, yet existing methods accept these unreliable suggestions without quality assessment.

\begin{figure}[t]
    \centering
    \includegraphics[width=0.98\columnwidth]{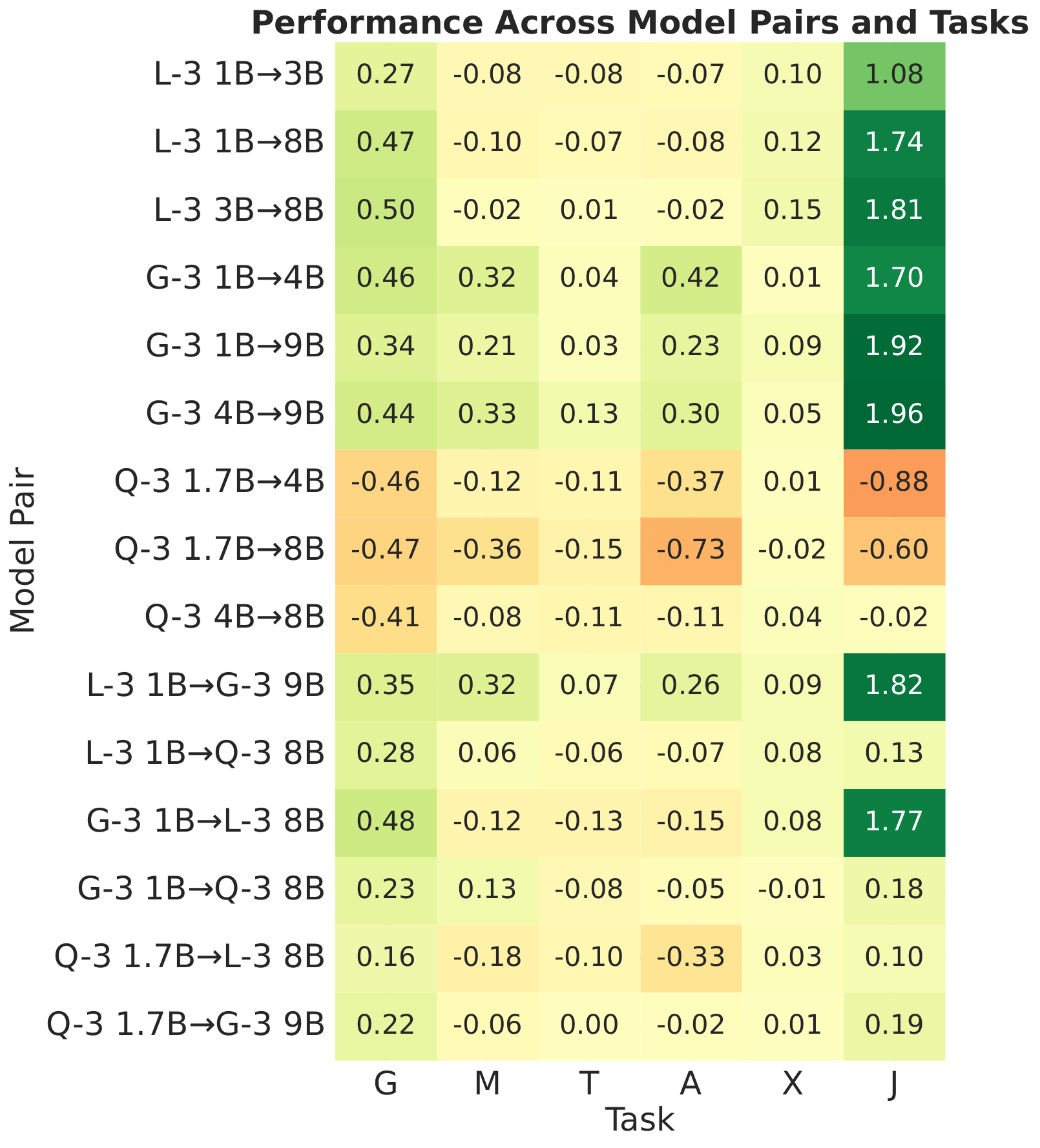}
    \caption{Difference between inference-time aligned performance and unaligned base model performance, showing if alignment has made performance better or worse. Red cells indicate degradation while green cells indicate improvement. Each model pair shows a guidance model aligning a base model by an arrow sign. L, G, and Q refer to Llama, Gemma and Qwen, while tasks are abbreviations of datasets GSM8K, MMLU, TruthfulQA, ARC-Challenge, XSTest and JustEval-Safe. Performance varies dramatically across guidance sources and benchmarks.}
    \label{fig:performance_heatmap}
    \vspace{-1.5em}
\end{figure}

\begin{figure*}[t]
    \centering
    \includegraphics[width=\textwidth]{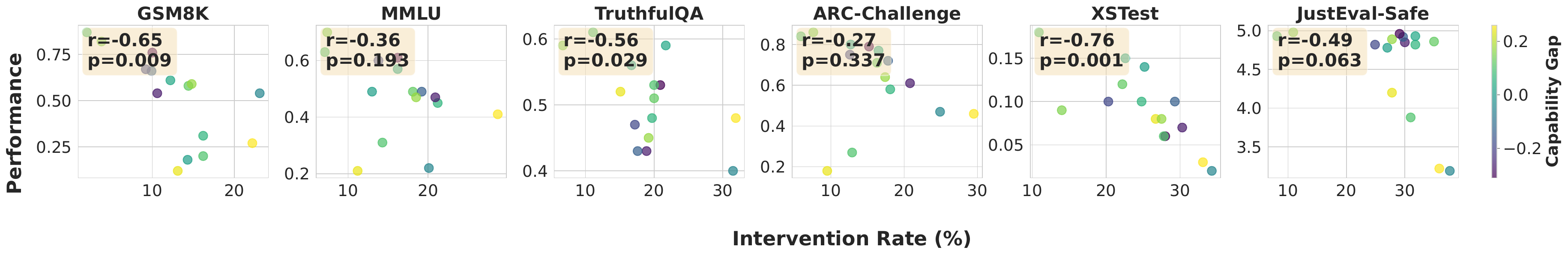}
    \caption{Intervention paradox: Higher intervention rates correlate with worse performance. Each point represents a model pair on a specific benchmark. Negative correlations are statistically significant on GSM8K, TruthfulQA, and XSTest. This contradicts the assumption that all guidance improves alignment. The paradox reveals that excessive intervention (>20\% of generated tokens in general, threshold may vary with the specific benchmark) actually degrades performance, 
    making intervention rate a diagnostic signal of 
    poor guidance quality.} 
    \label{fig:intervention_paradox}
    \vspace{-1.5em}
\end{figure*}


Figure~\ref{fig:performance_heatmap} presents our systematic evaluation on baseline alignment method~\cite{nudging} across nine models, 
three model families and six benchmarks. Performance for GSM8K~\cite{gsm8k}, MMLU~\cite{mmlu, mmlu2}, TruthfulQA~\cite{lin-etal-2022-truthfulqa}, ARC-Challenge~\cite{arcchallenge}, and XSTest~\cite{xstest} is measured by accuracy (0-1 scale), while performance for JustEval-Safe~\cite{justeval} is measured by average safety score (1-5 scale, higher is better). These 
metrics apply to all subsequent figures unless otherwise noted. 
Contrary to the implicit assumption that any
guidance is beneficial, we observe widespread failures that reveal critical insights. 
First, model family matters: When Qwen serves as the guidance model, its within-family pairs systematically fail across 
all benchmarks, while its cross-family pairs show a mix of successes and smaller failures. In contrast, Llama and Gemma demonstrate comparable successes both within and across families. 
This unpredictability greatly hinders practical deployment as we cannot reliably 
predict which model combinations may work without exhaustive testing.
Second, failures are not uniformly distributed. 
The same base model can have significant 
performance variance solely due to difference in guidance models. This suggests that guidance quality varies 
fundamentally across guidance models, yet existing methods lack mechanisms 
to account for these differences, creating a quality blindness problem.
These results motivate two questions that we later address: (1) Why do certain model 
pairs fail catastrophically? (2) Can we predict or mitigate failures without exhaustive testing?



\begin{figure*}[t]
    \centering
    \includegraphics[width=\textwidth]{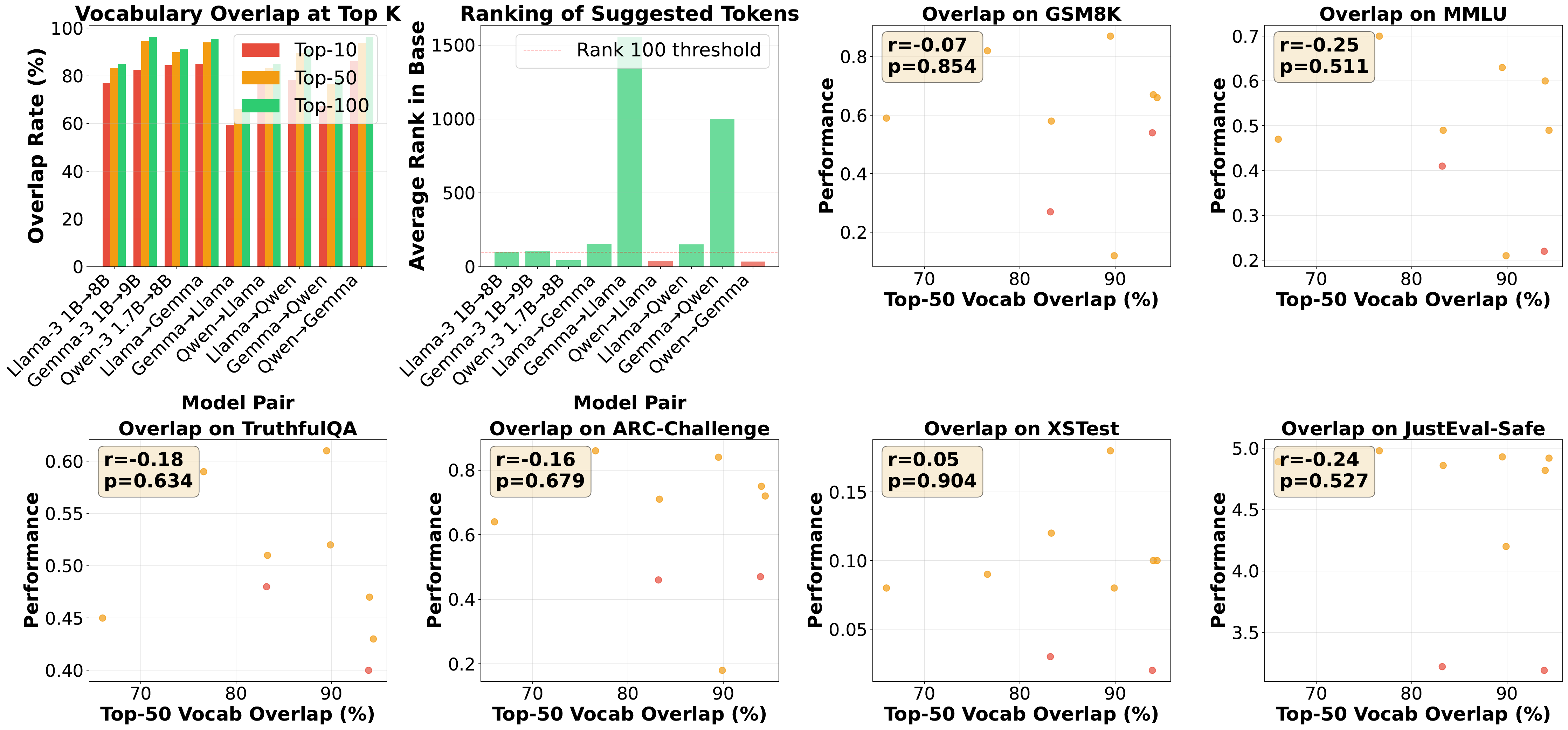}
    \caption{Vocabulary overlap does not predict performance. We measure top-50 vocabulary overlap, which is the percentage of suggested tokens from guidance model appearing in base model's top-50 predictions. Despite high overlap, pairs like Qwen→Llama could fail catastrophically, while low-overlap pairs like Gemma→Llama could succeed. No significant correlation between overlap and performance on any benchmark is found, rejecting the hypothesis that surface-level vocabulary alignment determines guidance quality.}
    \label{fig:vocabulary_analysis}
    \vspace{-1.5em}
\end{figure*}

Deeper empirical analysis reveals an intervention paradox that further confirms the quality blindness problem. As shown in Figure~\ref{fig:intervention_paradox}, intervention rate demonstrates a negative correlation with overall performance, statistically significant in benchmarks of GSM8K, TruthfulQA and XSTest. While 
this doesn't mean inference time alignment is not useful, 
excessive interventions (guided generation exceeds 20\% of the tokens generated) 
systematically 
show worse performance. 
This correlation implies that guidance models may as well struggle at positions where base model needs guidance and give misaligned or harmful suggestions. Such suboptimal suggestion creates 
misalignment 
in the generated sequence, introducing more problematic positions and hence more interventions. 
Existing methods have overlooked this and treat confident, helpful suggestions 
identically to misaligned, harmful ones, failing to detect 
or mitigate such problematic guidance. 
Further, this problem cannot be explained by superficial mechanisms, implying deeper significance. 
At first, we naively hypothesized that such failure is caused by tokenization mismatches. If the guidance model's top suggestion doesn't appear or ranks poorly in the base model's vocabulary distribution, guidance would be expected to become ineffective as the base model cannot continue coherently. 
We computed vocabulary overlap by generating 100 tokens under guidance and measured how often the guidance model's suggested token appears in the base model's top-k token candidates. 
We measure top-K overlap, which is the percentage of times where guidance model's suggestion is in base model's top-K. 
We also measure average rank, which is the mean rank of guidance model's suggestions in base model's distribution.  
We further computed Pearson correlation between vocabulary metrics and performance across all six benchmarks.
As shown in Figure~\ref{fig:vocabulary_analysis}, we found no statistically significant correlation between vocabulary overlap and overall performance. High-overlap pairs like Qwen-to-Llama can perform poorly, while low-overlap pairs like Gemma-to-Llama can perform well (See Figure~\ref{fig:performance_heatmap} for performance). 
This result rejects the naive hypothesis, suggesting deeper root cause and proving this issue non-trivial. 
Results using 90\% probability mass coverage are provided 
in Appendix~\ref{app:vocab_topp}, confirming the null 
finding is robust to overlap metric. 







\begin{figure*}[t]
    \centering
    \includegraphics[width=0.85\textwidth]{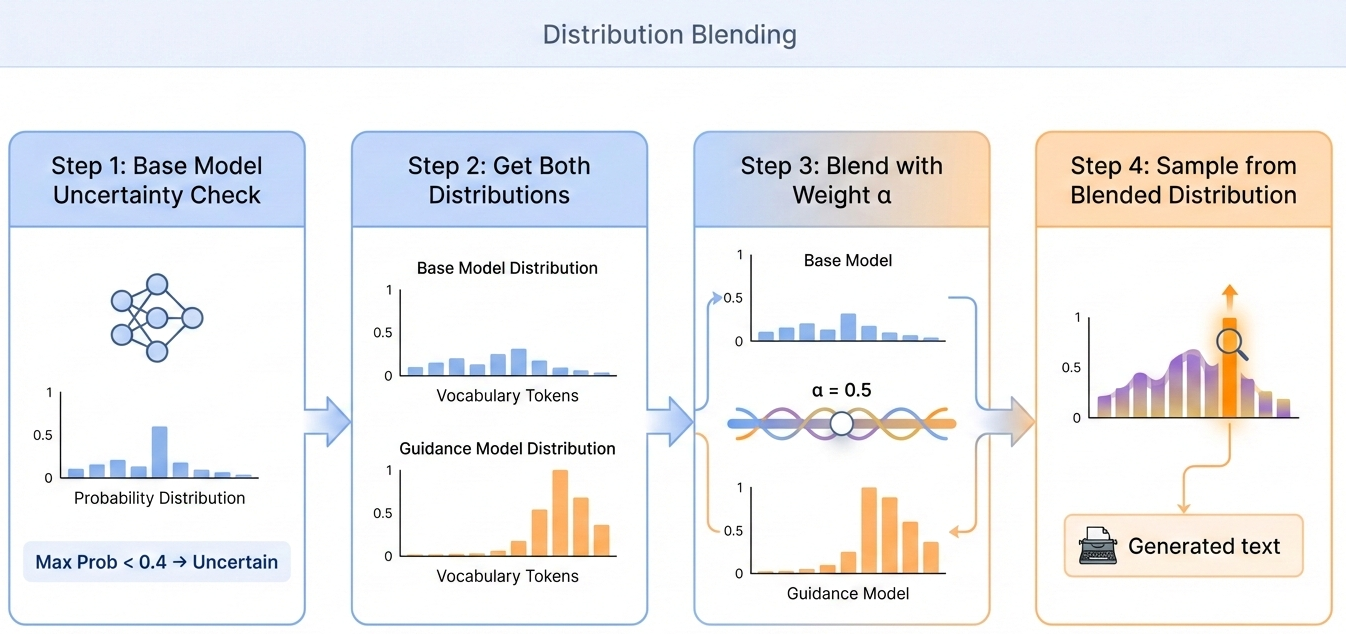}
    \caption{\name for Quality-Aware Inference-Time Alignment. 
    At positions where the base model is uncertain (max probability $< \tau$, default 0.4), 
    \name queries both base and guidance models for their full probability 
    distributions,  
    blends the distributions as $p_{\text{final}} = \alpha \cdot p_{\text{guidance}} 
    + (1-\alpha) \cdot p_{\text{base}}$, and 
    select the highest-probability token from the blended distribution (greedy decoding). 
    Full distributions are substitutable with top k to save computation, where k is an arbitrary large value. Unlike binary decisions in existing methods that either accepts or rejects suggestion, such soft integration preserves beneficial guidance while reducing 
    impact of unreliable suggestions, enabling quality-aware alignment.}
    \label{fig:method}
    \vspace{-1.5em}
\end{figure*}

\subsection{Our Solution} 
Given the intervention paradox, a natural question arises: since high intervention correlates with bad performance, can we simply cap 
intervention rate to improve alignment? We tested limiting intervention rate to 15\% by rejecting all guidance once the threshold is exceeded. We chose 15\% as a conservative threshold 
below the empirically observed 20\% to show even lenient limits 
degrade performance by discarding all guidance indiscriminately.

As Table~\ref{tab:capping} shows, intervention rate capping fails to address quality problems. Limiting interventions to 15\% degrades performance further on challenging pairs, as it removes both good and bad guidance indiscriminately. For guidance models with weak capability, removing bad guidance doesn't help if the base model truly needs assistance. For 
those 
with strong capability, capping removes beneficial guidance.
This validates that the problem is guidance quality rather than quantity. We need a selective mechanism to assess each suggestion's quality rather than 
a hard cap on 
intervention. High intervention rate is a symptom of guidance models making misaligned or harmful suggestions, not the cause of poor performance.

The intervention paradox exposes the fundamental gap of quality-assessment mechanisms. The null result of vocabulary analysis and the failure of naive solutions suggest the challenge of this problem. 
For solution, what's needed is not binary filtering but quality-aware integration that greedily selects candidate tokens from both models proportionally to their quality. 
This motivates our approach of \name, which we describe in Section~\ref{sec:method}. Meanwhile, the paradox reveals intervention rate as a diagnostic signal that predicts model incompatibility early from a small data subset rather than full benchmark evaluation. 



\section{\revjg{\name Method}}  \label{sec:method}

\subsection{Distribution Blending Method}

\name is designed as a stabilization technique for inference-time alignment: rather than replacing or overriding the base model's generation, it softly integrates guidance to prevent cascading failures. 

At each generation step $t$, given context $x_{<t}$:
\begin{itemize}[itemsep=2pt,topsep=2pt,parsep=2pt]
    \item Base model $M_b$ produces distribution $P_b(w | x_{<t})$ over vocabulary $\mathcal{V}$
    \item Guidance model $M_g$ produces distribution $P_g(w | x_{<t})$ over $\mathcal{V}$
    \item Base model's uncertainty: $u = \max_w P_b(w | x_{<t})$
\end{itemize}

When base is uncertain ($u < \tau$), integrate guidance in a quality-aware manner.



As illustrated in Figure~\ref{fig:method}, our approach consists of three steps at each generation position where base model is uncertain:

\textbf{Step 1: Obtain both distributions}

Query both models for their next-token distributions:
\begin{align}
\vspace{-1ex}
P_b(w | x_{<t}) &= \text{softmax}(\text{logits}_b(x_{<t})) \\
P_g(w | x_{<t}) &= \text{softmax}(\text{logits}_g(x_{<t}))
\end{align}

For computational efficiency, the full distribution could be substituted by only taking the top-k in distribution, where k is an arbitrary large value.

\textbf{Step 2: Compute blend weight}

Determine blending weight $\alpha \in [0,1]$ based on both models' confidence and token-level agreement:
\begin{equation}
\label{eq:alpha}
\alpha = \text{clip}\!\left(\frac{\hat{p}_g}{\hat{p}_b + \hat{p}_g} + \lambda \cdot P_b(t_g),\ 0,\ 1\right)
\end{equation}
where $\hat{p}_b = \max_w P_b(w|x_{<t})$ and $\hat{p}_g = \max_w P_g(w|x_{<t})$ are the top-1 probabilities of the base and guidance models respectively, $t_g = \arg\max_w P_g(w|x_{<t})$ is the guidance model's top token, and $\lambda{=}0.1$ controls the agreement bonus weight. The confidence ratio $\hat{p}_g / (\hat{p}_b + \hat{p}_g)$ is high when guidance is confident and base is uncertain, naturally scaling intervention strength. The agreement bonus $\lambda \cdot P_b(t_g)$ increases guidance influence when its top token already has support in the base distribution, reducing the risk of distributional mismatch. With $\lambda{=}0.1$, the bonus contributes at most a $0.1$ 
addition, 
keeping the confidence ratio as the primary driver. While this adaptive computation provides a principled default, $\alpha$ can also be manually tuned for task-specific optimal performance.
Practical hyperparameter tuning guidance is provided in Appendix~\ref{app:tuning}.

\textbf{Step 3: Blend and Greedy Selection}

Create a hybrid distribution that combines both models' knowledge, 
weighted by the blending weight $\alpha$. Each token $w$ in the 
vocabulary receives a probability that is a weighted average of 
the two models' predictions.
\begin{equation}
P_{\text{blend}}(w | x_{<t}) = \alpha \cdot P_g(w | x_{<t}) + (1-\alpha) \cdot P_b(w | x_{<t})
\end{equation}

We then select the token with the highest probability in the blended distribution:
\begin{equation}
\vspace{-1ex}
w_t = \arg\max_{w \in \mathcal{V}} P_{\text{blend}}(w | x_{<t})
\vspace{-1ex}
\end{equation}

Algorithm ~\ref{alg:blending} details a comprehensive overview of our algorithm. 
Sensitivity analysis for major hyperparameters is provided in Appendix~\ref{app:sensitivity}.

\begin{algorithm}[t]
\caption{Quality-Aware Distribution Blending}
\label{alg:blending}
\begin{algorithmic}[1]
\STATE \textbf{Input:} Base model $M_b$, guidance model $M_g$, prompt $x$
\STATE \textbf{Input:} Threshold $\tau$, blend weight $\alpha$ (see Eq.~\ref{eq:alpha} for adaptive default), max tokens $T$
\STATE \textbf{Output:} Sequence $y$
\STATE $y \leftarrow$ empty, $t \leftarrow 0$
\WHILE{$t < T$}
    \STATE Query $M_b$ for token probabilities $P_b$
    \STATE Compute $u \leftarrow$ maximum probability in $P_b$
    \IF{$u < \tau$}
        \STATE Query $M_g$ for token probabilities $P_g$
        \STATE For each token $w$: $P(w) \leftarrow \alpha P_g(w) + (1-\alpha) P_b(w)$
        \STATE Select token $w_t$ with highest $P(w_t)$
    \ELSE
        \STATE Select token $w_t$ with highest $P_b(w_t)$
    \ENDIF
    \STATE Append $w_t$ to $y$
    \STATE $t \leftarrow t + 1$
\ENDWHILE
\RETURN $y$
\end{algorithmic}
\end{algorithm}

\subsection{Why Distribution Blending Addresses Quality Blindness}

Rather than binary decisions, distribution blending allows both models to contribute partial information. Even 
guidance models that make misaligned or harmful predictions may have useful signal that, when properly downweighted, improves upon base model alone. 
Base model may also provide valuable information due to its greater capability. 
Blend weight could also adapt to base model's uncertainty level, in which more uncertain positions receive stronger guidance, while confident positions rely more on base, naturally reducing intervention when base has strong preferences. 
When guidance is misaligned or harmful, its contribution is downweighted to avoid 
catastrophic token selection, ensuring a graceful fallback. 

Cross-family blending operates on the shared tokens between models. We verified empirically that average overlap is >≈50\% across main datasets and representative model pairs, sufficient for meaningful distribution blending. Tokens are weighted by their respective model's contribution (α or 1-α) then renormalized,
ensuring 
coherent probability integration without requiring 
tokenizer alignment. 

\name also differs structurally from a conceptually-similar paradigm named confidence-based ensembling~\cite{CBE}. Confidence-based ensembling addresses the question of which model to trust more when aggregating predictions across multiple models. Confidence serves as a weight to resolve disagreement between models while maximizing prediction accuracy. Meanwhile, no model has directional authority over another. \name addresses a different problem: given a capable but unaligned base model and an aligned guidance model, how much alignment pressure to apply without destroying the base model’s capabilities? The blending is inherently directional, as we are pushing base distribution towards an aligned target. Here, confidence serves a different purpose from confidence-based ensembling: it regulates the degree of intervention on a per-sample basis to prevent capability degradation. This distinction matters because the two settings have different failure modes. In confidence-based ensembling, there is no directional pressure and models are aggregated without target distribution. The intervention paradox therefore cannot arise: it is a failure mode unique to alignment interventions, where pushing too hard toward a target distribution degrades model capability. This further highlights the novelty of our work.

\begin{table*}[t]
\centering
\setlength{\tabcolsep}{1.5pt}
\scriptsize
\resizebox{\linewidth}{!}{%
\begin{tabular}{l|cccccc|cccccc|cccccc}
\toprule
& \multicolumn{6}{c|}{GSM8K} 
& \multicolumn{6}{c|}{TruthfulQA} 
& \multicolumn{6}{c}{XSTest} \\
\cmidrule(lr){2-7}\cmidrule(lr){8-13}\cmidrule(lr){14-19}
Pair & Base & Guid. & Alig. & NUDG. & Ours & Int.\% 
     & Base & Guid. & Alig. & NUDG. & Ours & Int.\% 
     & Base & Guid. & Alig. & NUDG. & Ours & Int.\% \\
\midrule
\multicolumn{19}{c}{\textit{Cross-Family Pairs}} \\
\midrule
Q$\to$L & 0.11 & 0.53 & 0.87 & 0.27 & 0.31(+15\%) & 22.2 
         & 0.58 & 0.48 & 0.67 & 0.48 & 0.50(+4\%)  & 31.9 
         & 0.00 & 0.60 & 0.15 & 0.03 & 0.04(+33\%) & 33.1 \\
Q$\to$G & 0.32 & 0.53 & 0.86 & 0.54 & 0.56(+4\%)  & 23.1 
         & 0.40 & 0.48 & 0.70 & 0.40 & 0.42(+5\%)  & 31.5 
         & 0.01 & 0.60 & 0.14 & 0.02 & 0.02        & 34.3 \\
G$\to$L & 0.11 & 0.45 & 0.87 & 0.59 & 0.60(+2\%)  & 14.8 
         & 0.58 & 0.46 & 0.67 & 0.45 & 0.50(+11\%) & 19.2 
         & 0.00 & 0.96 & 0.15 & 0.08 & 0.08        & 27.5 \\
L$\to$G & 0.32 & 0.45 & 0.86 & 0.67 & 0.67        & 9.2  
         & 0.40 & 0.50 & 0.70 & 0.47 & 0.49(+4\%)  & 17.2 
         & 0.01 & 0.94 & 0.14 & 0.10 & 0.15(+50\%) & 20.3 \\
\midrule
\multicolumn{19}{c}{\textit{Within-Family Pairs}} \\
\midrule
L$\to$L & 0.11 & 0.45 & 0.87 & 0.58 & 0.58        & 14.4 
         & 0.58 & 0.50 & 0.67 & 0.51 & 0.51        & 20.0 
         & 0.00 & 0.94 & 0.15 & 0.12 & 0.12        & 22.2 \\
G$\to$G & 0.32 & 0.45 & 0.86 & 0.66 & 0.66        & 10.0 
         & 0.40 & 0.46 & 0.70 & 0.43 & 0.43        & 20.9 
         & 0.01 & 0.96 & 0.14 & 0.10 & 0.14(+40\%) & 28.0 \\
\bottomrule
\end{tabular}
}
\caption{Performance of unaligned base model (Base), standalone guidance model (Guid.), base model aligned by training (Alig.), NUDGING baseline (NUDG.), and our method (Ours). We also show NUDGING intervention rates (Int.\%). Abbreviations: Q=Qwen, L=Llama, G=Gemma. Percentages show improvement over NUDGING. Guidance models mostly outperform unaligned base models, 
demonstrating their capability to guide the base models. Even when base models outperform guidance models in some scenarios of TruthfulQA, our method has already demonstrated improvements compared with unaligned base models, indicating successful alignment.
Overall, 
our method achieves improvements on high-intervention pairs while maintaining performance on low-intervention pairs.}
\label{tab:main_results}
\vspace{-2em}
\end{table*}

\section{Experiments}

We evaluate nine recently-released models of different sizes within three families: Llama~\cite{llama3}, Gemma~\cite{gemma3, gemma2} and Qwen~\cite{qwen3}. They are tested in pairs both within and across family. 
We test on benchmarks like GSM8K~\cite{gsm8k}, TruthfulQA~\cite{lin-etal-2022-truthfulqa} and XSTest~\cite{xstest} where the intervention paradox is statistically 
significant (Figure~\ref{fig:intervention_paradox}). For reasoning tasks, GSM8K contains grade-school math word problems and TruthfulQA measures factual accuracy. For safety tasks, XSTest consists of adversarial safety prompts. 
We report results on a fixed random subset of 100 samples per benchmark to balance computational cost with statistical reliability. We also use greedy decoding (temperature=0.0), which produces deterministic 
outputs for reproducibility. 


We use the default $\tau = 0.4$ as the threshold for base model uncertainty,  triggering guidance 
when $\max_w P_b(w | x_{<t}) < 0.4$. Other hyperparameters are kept default. 

We measure task performance as accuracy (percentage of correct answers) and also report 
intervention rate (percentage of tokens where guidance occurs) of baseline.

For \name, we obtain top-100 token probabilities from 
both models and blend them. 
We evaluate on representative cross-family pairs spanning the performance spectrum: high-intervention poor-performing cases (Qwen-to-Llama/Gemma) and low-intervention well-performing cases (Gemma-Llama pairs). We also include well-performing within-family pairs like Gemma and Llama. 

We focus on the practical setting where small guidance models guide larger 
base models to minimize computational costs. 
Guidance models include: Llama-3.2-1B-Instruct, Gemma-3-1b-it, Qwen3-1.7B. 
Base models include: Llama-3.1-8B, Gemma-2-9b, Qwen3-8B-Base. 






We compare against the following baselines: 

\textbf{Base Model:} Base model generates alone without guidance.

\textbf{Guidance Model:} Guidance model generates alone to prove qualification for guidance. 

\textbf{Aligned base model}: Base model aligned through finetuning, establishing 
the upper bound for alignment effectiveness. 

\textbf{NUDGING}~\cite{nudging}: 
At uncertain positions (max prob $< \tau$), base model accepts guidance model's suggestions. 

\textbf{Intervention Capping:} Naive solution to address intervention 
paradox by rejecting all guidance once intervention rate exceeds threshold 
(15\%). We include this to demonstrate that simple quantity-based approaches fail.


Table~\ref{tab:main_results} presents our main results. 
\name achieves consistent improvements on high-intervention pairs. 
Qwen-guided pairs, which exhibit consistently high intervention rates across all tasks (22-34\%), show the most substantial improvements. On Qwen→Llama, we achieve +15\% on GSM8K (0.27→0.31), +4\% on TruthfulQA (0.48→0.50), and +33\% on XSTest (0.03→0.04). Qwen→Gemma shows similar patterns with +4-5\% improvements on GSM8K and TruthfulQA. 
These results validate 
that our method 
addresses quality failures most prevalent in high-intervention scenarios, where guidance sources intervene frequently and provide unreliable suggestions. 

Improvements on other cross-family pairs are task-dependent. 
Gemma→Llama and Llama→Gemma demonstrate various intervention rates (9-28\%) and alignment successes (+2-50\%) depending on task difficulty. 
Notably, Gemma→Llama achieves +11\% on TruthfulQA despite only 19\% intervention rate, while Llama→Gemma shows +50\% on XSTest at 20\% intervention. This variability suggests that intervention rate alone does not determine improvement potential. 
Improvements also vary by task because guidance quality is task-dependent: 
Gemma provides better guidance on TruthfulQA while Llama provides better 
guidance on XSTest, and our method leverages this selectively. 

\name also preserved performance on low-intervention pairs. 
Within-family pairs (Llama→Llama, Gemma→Gemma) maintain baseline performance across most configurations, with intervention rates ranging 10-28\%. The one notable exception is Gemma→Gemma on XSTest (+40\%, 0.10→0.14), where high intervention (28\%) on this safety task benefits from \name. Critically, we observe no catastrophic degradation: even when our method provides no improvement, performance remains stable, demonstrating robustness across the intervention spectrum.



We further validate on full test sets with three representative model 
pairs spanning the intervention spectrum: Qwen$\to$Llama and 
Qwen$\to$Gemma (high intervention rate, poor baseline performance) and Llama$\to$Gemma 
(low intervention, good baseline performance). 
As shown in Table~\ref{tab:full_results}, 
while individual improvements are modest (+0--4\%) and do not 
consistently reach statistical significance, the pattern is consistent: 
our method shows improvements on high-intervention scenarios (>31\%) while 
preserving performance in low-intervention scenarios. Results can be further 
optimized by task-specific $\alpha$ tuning. 
This task-dependent pattern also aligns with intuition: large base models 
retain strong reasoning capability despite lacking alignment, requiring 
less guidance on GSM8K. However, factual objectivity (TruthfulQA) and 
safety (XSTest) are precisely where alignment deficits manifest and 
intervention provides most benefit. 
Note that these results are under 1B guidance models which are limited in capability--- we reasonably expect greater improvement if using larger models.

Overall, our approach achieves improvements in high-intervention scenarios while maintaining good performance in low-intervention cases where 
guidance quality is already adequate. 
This shows that \name addresses the fundamental quality assessment gap in inference-time alignment. 
While inference-time methods cannot fully replace 
finetuning alignment yet in effectiveness, they provide significant value where retraining is 
prohibitive.

\begin{table}[t]
\centering
\small
\setlength{\tabcolsep}{3pt}
\begin{tabular}{l|l|cc|c}
\toprule
Dataset & Pair & Int.\% & NUDGING & Ours \\
\midrule
\multirow{3}{*}{GSM8K}
  & Q$\to$L & 22.4 & $0.29_{\pm.024}$ & $0.29_{\pm.024}$ \\
  & L$\to$G & 10.0 & $0.59_{\pm.027}$ & $0.59_{\pm.027}$ \\
  & Q$\to$G & 23.5 & $0.50_{\pm.027}$ & $0.50_{\pm.027}$ \\
\midrule
\multirow{3}{*}{TruthfulQA}
  & Q$\to$L & 31.5 & $0.46_{\pm.034}$ & $0.47_{\pm.034}$ \\
  & L$\to$G & 16.6 & $0.48_{\pm.034}$ & $0.48_{\pm.034}$ \\
  & Q$\to$G & 30.9 & $0.44_{\pm.034}$ & $0.44_{\pm.034}$ \\
\midrule
\multirow{3}{*}{XSTest}
  & Q$\to$L & 35.3 & $0.52_{\pm.062}$ & $0.54_{\pm.062}$ \\
  & L$\to$G & 22.9 & $0.90_{\pm.037}$ & $0.90_{\pm.037}$ \\
  & Q$\to$G & 35.4 & $0.47_{\pm.062}$ & $0.48_{\pm.062}$ \\
\bottomrule
\end{tabular}
\caption{Full test set evaluation with 95\% confidence intervals. 
While individual improvements are modest (+0--4\%) 
and do not consistently reach statistical significance, 
the pattern is consistent: 
our method shows improvements on high-intervention scenarios while 
preserving performance in low-intervention scenarios, 
validating the effectiveness of our 
approach. The modest magnitude also underscores our recommendation 
for task-specific $\alpha$ optimization.}
\label{tab:full_results}
\vspace{-1em}
\end{table}

\begin{table}[t]
\centering
\small
\setlength{\tabcolsep}{4pt}  
\begin{tabular}{lcccc}
\toprule
Method & Q→L & Q→G & G→L & L→G \\
& (22\%) & (23\%) & (15\%) & (9\%) \\
\midrule
NUDGING & 0.27 & 0.54 & 0.59 & 0.67 \\
Cap at 15\% & 0.25 & 0.46 & 0.55 & 0.58 \\
\midrule
Ours & \textbf{0.31} & \textbf{0.56} & \textbf{0.60} & \textbf{0.67} \\
\bottomrule
\end{tabular}
\caption{Intervention rate capping degrades performance across all pairs on GSM8K 
by discarding guidance indiscriminately. Numbers in parentheses show 
baseline intervention rates. }
\label{tab:capping}
\vspace{-2em}
\end{table}

Table~\ref{tab:capping} demonstrates that intervention rate capping, which is a 
naive way to address the intervention paradox by limiting guidance 
acceptance to 15\%, consistently degrades performance. This approach fails 
because it removes all guidance after reaching the threshold, 
discarding both beneficial and harmful interventions. The degradation 
affects all pairs regardless of baseline intervention rate: high-intervention 
pairs (Qwen→Gemma: 0.54→0.46, -15\%), moderate pairs (Gemma→Llama: 
0.59→0.55, -7\%), and even low-intervention pairs already below the 15\% 
threshold (Llama→Gemma: 0.67→0.58, -13\%). 
This validates the necessity of selectively filtering harmful 
interventions while preserving beneficial ones instead of hard-limiting 
intervention frequency.
We further evaluate discrete filtering alternatives in 
Appendix~\ref{app:filtering}, which confirm that 
acceptance-rule-based approaches are insufficient.

\section{Conclusion}

Inference-time alignment methods could transfer alignment properties 
from aligned LLMs to unaligned LLMs without retraining. 
Through comprehensive analysis across nine models and six benchmarks, we 
make novel contributions toward systematic 
characterization of quality failures in inference-time 
alignment. 
We identify the intervention paradox: excessive intervention rates correlate with 
worse performance, 
contradicting the assumption that all 
guidance improves alignment. This paradox reveals that existing methods 
suffer from quality blindness as they lack guidance quality assessment mechanisms to 
distinguish beneficial guidance from harmful ones.

To address quality blindness, we propose \name, which softly 
integrates both models' probability distributions at difficult positions 
rather than making binary accept-or-reject decisions. 
This quality-aware 
approach achieves consistent improvements on high-intervention pairs 
while 
maintaining performance on low-intervention pairs. 
Overall, our work establishes diagnostic 
signals for rapid failure detection and demonstrates that quality-aware 
integration enables more robust cross-model guidance. This foundation opens 
pathways for developing better inference-time methods that reliably improve model 
performance across diverse model combinations.

\section*{Limitations} 
Our work focuses on characterizing and mitigating quality failures but 
does not provide predictive models for determining compatibility before 
testing. The intervention rate serves as a rapid diagnostic signal 
measurable in a small subset, but deployment still requires validating each 
model pair empirically. Additionally, while \name outperforms baselines, improvements on some pairs remain 
modest, 
suggesting room for further optimization through hyperparameter tuning,  
adaptive blending strategies or representation-level compatibility 
assessment.

\section*{Acknowledgments}
This research is supported by MOE Tier 1 grant RG16/22.

\vspace{-1em}
\bibliography{custom}

\appendix

\section{Appendix}
\label{sec:appendix}

\subsection{Vocabulary Overlap under Top-\textit{p} Coverage}
\label{app:vocab_topp}
We also computed overlap using 90\% probability mass coverage for further validation. Results in Table ~\ref{tab:vocab_topp} confirm the null finding across all six benchmarks (|r| ≤ 0.35, p > 0.35), with top-p(0.9) yielding smaller or equivalent overlap sets compared with top-50 in 8 of 9 model pairs (three Within-family + six Cross-family). The stricter metric does not reveal any hidden correlation.

These results exactly explain why we originally didn’t consider probability mass coverage: top-k=50 already provides the most favorable conditions for detecting a correlation. Top-k=50 is deliberately lenient: it includes many low-probability tokens, maximizing the overlap set size and thus maximizing statistical power to detect any relationship between vocabulary overlap and generation quality. If guidance tokens appearing in this large, inclusive set show no predictive power (r≤0.25, p>0.05), then a stricter 90\% probability mass criterion that selects only the model's highest-confidence tokens and yields smaller overlap sets would have less power to detect such a relationship. The null result under our favorable conditions is therefore informative.

\begin{table}[t]
\centering
\small
\begin{tabular}{l|rr|rr}
\toprule
& \multicolumn{2}{c|}{Top-$k$ ($k{=}50$)} 
& \multicolumn{2}{c}{Top-$p$ ($p{=}0.9$)} \\
Benchmark & \multicolumn{1}{c}{$r$} & \multicolumn{1}{c}{$p$} & \multicolumn{1}{c}{$r$} & \multicolumn{1}{c}{$p$} \\
\midrule
GSM8K          & $-$0.072 & 0.853 & $-$0.170 & 0.662 \\
MMLU           & $-$0.253 & 0.511 & $-$0.351 & 0.355 \\
TruthfulQA     & $-$0.186 & 0.632 & $-$0.336 & 0.376 \\
ARC-Challenge  & $-$0.161 & 0.679 & $-$0.233 & 0.547 \\
XSTest         & $+$0.047 & 0.905 & $-$0.073 & 0.851 \\
JustEval-Safe  & $-$0.244 & 0.527 & $-$0.326 & 0.392 \\
\bottomrule
\end{tabular}
\caption{Pearson correlation between vocabulary overlap and performance under top-$k$ ($k{=}50$) and top-$p$ ($p{=}0.9$) coverage metrics. Neither metric reveals a statistically significant correlation ($|r| \leq 0.35$, $p > 0.35$ in all cases). Top-$p$(0.9) yields smaller or equivalent overlap sets in 8 of 9 model pairs yet does not uncover any hidden relationship, confirming that the null finding is robust to overlap metric choice.}
\label{tab:vocab_topp}
\end{table}

\subsection{Hyperparameter Sensitivity}
\label{app:sensitivity}

\begin{table}[t]
\centering
\small
\begin{tabular}{l|ccc}
\toprule
$\tau$ & GSM8K & TruthfulQA & XSTest \\
\midrule
0.3 & 0.25 & 0.42 & 0.50 \\
0.4 & 0.12 & 0.30 & 0.52 \\
0.5 & 0.32 & 0.44 & 0.55 \\
\bottomrule
\end{tabular}
\hfill
\begin{tabular}{l|ccc}
\toprule
$\alpha$ & GSM8K & TruthfulQA & XSTest \\
\midrule
0.3  & 0.23 & 0.47 & 0.45 \\
0.5  & 0.28 & 0.42 & 0.52 \\
0.7  & 0.12 & 0.42 & 0.54 \\
Auto & 0.12 & 0.30 & 0.52 \\
\bottomrule
\end{tabular}
\caption{Sensitivity of $\tau$ with $\alpha{=}$Auto (\textit{left}) and $\alpha$ with $\tau{=}0.4$ (\textit{right}) on Qwen$\to$Llama (full test set). Both hyperparameters show task-dependent effects, empirically validating the recommendation of task-specific manual tuning. Default settings ($\tau{=}0.4$, $\alpha{=}$Auto) provide reasonable bottomline performances across tasks without tuning.}
\label{tab:sensitivity}
\vspace{-1.5em}
\end{table}

We also conducted sensitivity analysis for τ (uncertainty threshold, established and validated in prior work on inference-time alignment~\cite{nudging}) and α (blend weight, one of our contributions) on representative challenging model pair Qwen->Llama and full datasets. As shown in table~\ref{tab:sensitivity}, both τ and α show task-dependent effects. This empirically validates our stated recommendation that α should be manually tuned for task-specific optimal performance. Similarly, τ controls intervention frequency and benefits from task-specific tuning. The observed variance reflects that different tasks benefit from different hyperparameter settings.

\subsection{Discrete Filtering Baselines}
\label{app:filtering}

\begin{table}[t]
\centering
\footnotesize
\setlength{\tabcolsep}{2pt}
\begin{tabular}{l|rrr|rrr|rrr}
\toprule
\multicolumn{1}{c|}{Dataset} & \multicolumn{3}{c|}{GSM8K}
& \multicolumn{3}{c|}{TruthfulQA} 
& \multicolumn{3}{c}{XSTest} \\
\cmidrule(lr){2-4}\cmidrule(lr){5-7}\cmidrule(lr){8-10}
\multicolumn{1}{c|}{Method} & LL & GG & QL
       & LL & GG & QL 
       & LL & GG & QL \\
\midrule
NUDG.      & 0.58 & 0.66 & 0.27 
             & 0.51 & 0.43 & 0.43 
             & 0.12 & 0.10 & 0.03 \\
AF & \multirow{2}{*}{0.45} 
             & \multirow{2}{*}{0.41} 
             & \multirow{2}{*}{0.28}
             & \multirow{2}{*}{0.36} 
             & \multirow{2}{*}{0.28} 
             & \multirow{2}{*}{0.22}
             & \multirow{2}{*}{0.10} 
             & \multirow{2}{*}{0.12} 
             & \multirow{2}{*}{0.03} \\
CC & & & & & & & & & \\
\midrule
Ours         & \textbf{0.58} & \textbf{0.66} & \textbf{0.31} 
             & \textbf{0.51} & \textbf{0.43} & \textbf{0.50} 
             & \textbf{0.12} & \textbf{0.10} & \textbf{0.04} \\
\bottomrule
\end{tabular}
\caption{Comparison against the discrete filtering baseline combining Agreement Filter and Confidence Competition. 
Pair abbreviations: LL=Llama$\to$Llama, GG=Gemma$\to$Gemma, 
QL=Qwen$\to$Llama. Agreement Filter (AF) accepts guidance 
only when the guidance token appears in the base model's top-$k$; 
Confidence Competition (CC) accepts guidance only when 
guidance model top-1 probability exceeds the base model's. The baseline shows inconsistent and mostly worse results than NUDGING, 
confirming that soft quality-aware integration is necessary. Our approach maintains or improves over NUDGING across all 
configurations.}
\label{tab:filtering}
\vspace{-1.5em}
\end{table}

We additionally evaluated the discrete filtering baseline combining Agreement Filter and Confidence Competition. Agreement Filter accepts guidance only when the guidance token appears in the 
base model's top-$k$ candidates; Confidence Competition accepts 
guidance only when the guidance model's top-1 probability exceeds 
the base model. As shown in Table~\ref{tab:filtering}, the baseline 
performs inconsistently and mostly worse than main baseline Nudging, confirming that 
the intervention paradox cannot be addressed by discrete acceptance 
rules and motivating our soft blending approach.

\subsection{Practical Hyperparameter Tuning Guide}
\label{app:tuning}
For deployment, we recommend the following lightweight protocol:

\textbf{Default settings:} Use $\tau$=0.4 (base uncertainty threshold) and 
$\alpha$=auto (adaptive blending). These provide reasonable performance without 
tuning.

\textbf{When to tune:} If initial results show intervention rate exceeding 20\% with unsatisfactory 
performance, optimize hyperparameters on a small 
validation set (100 samples).

\textbf{Tuning protocol:}
\vspace{-.5em}
\begin{enumerate}[itemsep=0pt]
\item Fix $\tau$=0.4, test $\alpha \in \{0.3, 0.5, 0.7\}$ 
\item Optionally refine $\alpha$ by ±0.1 around best value so far for further improvement on performance 
\item Select best $\alpha$, then test $\tau \in \{0.3, 0.4, 0.5\}$ 
\end{enumerate}

Each configuration requires around 5 minutes on 100 samples, making 
basic protocols (steps 1 and 3, 6 trials in total) 
feasible within 30 minutes. 
Coarse-grained tuning (steps 1-3) captures most potential gains, while finer adjustments 
(e.g., α ±0.05) 
yield diminishing returns relative to validation cost.

\end{document}